\documentclass[preprint,12pt,authoryear]{elsarticle}

\usepackage{amssymb}
\usepackage{amsmath}
\usepackage{csquotes}
\usepackage{amsthm}
\usepackage[outdir=./]{epstopdf}
\usepackage{hyperref}

\usepackage{lineno}
\usepackage{booktabs} % For better table formatting
\usepackage{multirow} % For multirow feature in tables
\usepackage[table]{xcolor}  % Allows for row coloring
\usepackage{graphicx}

\begin{document}

\begin{frontmatter}

%% Title, authors and addresses

%% use the tnoteref command within \title for footnotes;
%% use the tnotetext command for theassociated footnote;
%% use the fnref command within \author or \affiliation for footnotes;
%% use the fntext command for theassociated footnote;
%% use the corref command within \author for corresponding author footnotes;
%% use the cortext command for theassociated footnote;
%% use the ead command for the email address,
%% and the form \ead[url] for the home page:
%% \title{Title\tnoteref{label1}}
%% \tnotetext[label1]{}
%% \author{Name\corref{cor1}\fnref{label2}}
%% \ead{email address}
%% \ead[url]{home page}
%% \fntext[label2]{}
%% \cortext[cor1]{}
%% \affiliation{organization={},
%%            addressline={}, 
%%            city={},
%%            postcode={}, 
%%            state={},
%%            country={}}
%% \fntext[label3]{}

\title{AI Algorithm for Predicting and Optimizing Trajectory of UAV Swarm}

%% use optional labels to link authors explicitly to addresses:
%% \author[label1,label2]{}
%% \affiliation[label1]{organization={},
%%             addressline={},
%%             city={},
%%             postcode={},
%%             state={},
%%             country={}}
%%
%% \affiliation[label2]{organization={},
%%             addressline={},
%%             city={},
%%             postcode={},
%%             state={},
%%             country={}}

\author{Amit Raj\fnref{label1}}

\author{Kapil Ahuja\fnref{label1}\corref{cor1}}

\cortext[cor1]{Corresponding author:}
\ead{kahuja@iiti.ac.in}

\affiliation[label1]{organization={Math of Data Science \& Simulation (MODSS) Lab, Computer Science \& Engineering},%Department and Organization
            addressline={IIT Indore}, 
            city={Indore},
            postcode={453552},   
            state={Madhya Pradesh},
            country={India}}

\author{Yann Busnel\fnref{label3}}

\affiliation[label3]{organization={IMT Nord Europe – Campus de Recherche},%Department and Organization
            city={Douai},
            postcode={F-59508}, 
            state={Nord},
            country={France}}

\begin{abstract}

This paper explores the application of Artificial Intelligence (AI) techniques for generating the trajectories of fleets of Unmanned Aerial Vehicles (UAVs). The two main challenges addressed include accurately predicting the paths of UAVs and efficiently avoiding collisions between them.

{\it Firstly}, the paper systematically applies a diverse set of activation functions to a Feedforward Neural Network (FFNN) with a single hidden layer, which enhances the accuracy of the predicted path compared to previous work. 
\par
{\it Secondly}, we introduce a novel activation function, AdaptoSwelliGauss, which is a sophisticated fusion of Swish and Elliott activations, seamlessly integrated with a scaled and shifted Gaussian component. Swish facilitates smooth transitions, Elliott captures abrupt trajectory changes, and the scaled and shifted Gaussian enhances robustness against noise. This dynamic combination is specifically designed to excel in capturing the complexities of UAV trajectory prediction. This new activation function gives substantially better accuracy than all existing activation functions.
\par
{\it Thirdly}, we propose a novel Integrated Collision Detection, Avoidance, and Batching (ICDAB) strategy that merges two complementary UAV collision avoidance techniques: changing UAV trajectories and altering their starting times, also referred to as batching. This integration helps overcome the disadvantages of both—reduction in the number of trajectory manipulations, which avoids overly convoluted paths in the first technique, and smaller batch sizes, which reduce overall takeoff time in the second.

\end{abstract}

% %%Graphical abstract
% \begin{graphicalabstract}
% %\includegraphics{grabs}
% \end{graphicalabstract}

% %%Research highlights
% \begin{highlights}
% \item Research highlight 1
% \item Research highlight 2
% \end{highlights}


\end{frontmatter}

% \begin{IEEEkeywords}
% Trajectory planning, Autonomous flight control, Deep Learning, Multi-UAV coordination
% \end{IEEEkeywords}
\section{Introduction}
UAVs have become increasingly popular in recent years due to their versatility and potential for a wide range of applications, from surveillance and monitoring to delivery and transportation. However, safe and efficient operation of UAVs in complex environments remains a significant challenge, particularly when multiple UAVs are involved. A key issue is the need to optimize the trajectories of the UAVs to achieve various objectives, such as minimizing travel time, avoiding collisions, and maximizing coverage. Traditional methods for trajectory planning and control are often limited in their ability to handle the complexity and uncertainty of real-world scenarios and may not be scalable to large fleets of UAVs.

Prior research, exemplified by \cite{lai2020machine}, \cite{xue2017uav}, and \cite{qiu2020multi}, has demonstrated the efficacy of leveraging non-linear optimization techniques. Recently, \cite{xu2024multi} have used multi-objective optimization for trajectory generation. When quick trajectory changes are required, the optimization routine is too slow and not adaptive, and hence, AI techniques are preferred. AI techniques, particularly those based on machine learning and neural networks, have shown great promise in addressing these challenges by enabling UAVs to learn from data and adapt to changing conditions \cite{lai2020machine}. 

These studies involved training FFNN with a single hidden layer, utilizing activation functions such as the Hyperbolic Tangent function (Tanh), Sigmoid, etc. The activation functions used do not predict the path with much accuracy, and hence, we {\it first} improve this aspect. We systematically apply a diverse set of activation functions to an FFNN with a single hidden layer, conducting a comprehensive comparative analysis. Besides Sigmoid and Tanh, we use  Rectified Linear Unit (ReLU), Leaky ReLU, Swish, Elliot, and Maxout.

In our pursuit of enhanced trajectory accuracy, {\it second}, we introduce a novel activation function, AdaptoSwelliGauss, which surpasses commonly used counterparts in the same neural network architecture. This function combines the Swish activation function, which captures smooth transitions and maintains trajectory continuity, with the Elliot activation function, which captures abrupt shifts in direction and velocity, as well as a scaled and shifted Gaussian, which makes the activation function robust against noisy data.

In autonomous UAVs, the crucial components of collision detection and avoidance play a paramount role in ensuring the safety and efficiency of their operations, which is our {\it third} focus. The significance of these features becomes even more pronounced when considering multiple UAVs taking off simultaneously. Collision detection between UAVs is straightforward, however, there are many ways to avoid collisions by changing their trajectories. \cite{guo2021three} introduces one such popular approach, the Circular Arc Trajectory Geometric Method (CTGA). Assuming two UAVs are colliding at a point in the path, this technique adds a small perturbation to the path of one of these UAVs. A drawback of this algorithm is its susceptibility to getting stuck in a manipulation loop. Any alteration in the trajectory of one UAV may inadvertently create collision candidates with other UAVs, leading to a challenging situation. Additionally, frequent manipulations in a UAV's trajectory can result in a convoluted flight path, compromising the overall efficiency of the UAV swarm.

Another complementary technique to avoid UAV collisions is to change their starting times. \cite{sastre2022collision} and \cite{sastre2022safe} propose such a popular approach. They employ a batching mechanism, creating groups of UAVs with non-colliding trajectories to facilitate safe flight. However, the creation of multiple batches introduces a time-intensive process, delaying the overall launch of the UAV swarm.

In this paper, we introduce an advanced collision detection and avoidance algorithm, referred to as the ICDAB algorithm. Here, we first improve the CTGA algorithm from \cite{guo2021three}, and then we integrate this avoidance algorithm with the batching mechanism leading to our algorithm.

The remainder of the paper is organized as follows: Section \ref{Literature} reviews the literature, Section \ref{Methodology} describes our proposed algorithms and methodology, Section \ref{Result} presents the results, and Section \ref{Conclusion} concludes the paper and suggests directions for future work.

\section{Literature Review} \label{Literature}

In this section, we analyze the previous works on applying AI algorithms in optimizing UAV trajectories, and their collision detection and avoidance.

{\it Firstly}, we highlight the earlier studies on AI for UAVs, as summarized in Table~\ref{tab:ai_UAVs}. As evident, these papers address different kinds of problems. Papers \cite{xue2017uav}, \cite{xiao2019trajectory} and \cite{lai2020machine} are closest to our problem. The activation functions used by them are standard, archiving errors (Mean Squared Error or MSE and Root MSE or RMSE) in the order of \(10^{-3}\). However, we experiment with a large number of activation functions (including our new activation function), achieving error in the order of the order \(10^{-11}\) to \(10^{-14}\).
\begin{table}[htbp]
\centering
\fontsize{9pt}{12pt}\selectfont
\caption{Summary on AI for UAVs.}
\vspace{2mm}
\label{tab:ai_UAVs}
\begin{tabular}{ p{2.4cm} p{2.5cm} p{2.5cm} p{2cm}  p{2.2cm}} 
\hline
Studies & Focus & AI Architecture &  {Activation Function} & Error \\
\hline
\cite{xue2017uav} & Trajectory Modelling & FFNN & Tanh & MSE: 0.003\\
\hline
\cite{xiao2019trajectory} & Predict UAV's Position   & Recurrent Neural Network & Tanh  & MSE: 0.005\\
\hline
\cite{lai2020machine}& Trajectory Generation & FFNN  & Tanh & RMSE: 0.170\\
\hline
\cite{sarkar2020intelligent} & Flight Time Prediction & FFNN & Tanh & MSE: 0.0016\\
\hline
\cite{jeong2021hazardous} & Wind-induced Trajectory Deviation &Deep Neural Network (DNN12) & ReLU & RMSE: 0.007\\
\hline
\cite{jiang2022neural} & UAV Control & FFNN  & Sigmoid & RMSE: 0.126\\
\hline
\end{tabular}
\end{table}

{\it Second}, we review literature on collision detection and avoidance, specifically trajectory manipulation during flight in Table ~\ref{tab:ca}. The most common technique used is detailed in \cite{guo2021three}. As mentioned earlier, we further improve this algorithm and integrate it with a collision avoidance technique involving changes to the UAVs' start times (or batching), which is discussed in the next table.

\begin{table}[htbp]
\centering
\caption{Summary on UAV Swarm Collision Avoidance.}
\vspace{2mm}
\fontsize{9pt}{12pt}\selectfont
\label{tab:ca}
\begin{tabular}{p{3.5cm} p{4.5cm} p{4.5cm}}
\hline
Studies  & Focus & Key Techniques/Methods Employed \\
\hline
\cite{lin2017sampling} & Collision-free path planning & Closed-loop rapidly-exploring random tree algorithm \\
\hline
\cite{elmkaiel2019collision} &  Quadcopter swarm collision avoidance algorithm & Algorithm based on repulsive force fields \\
\hline
\cite{wan2019distributed} &  Distributed conflict detection and resolution  & Consensus and leader-follower strategy algorithms\\
\hline
\cite{han2019efficient} & Efficient collision-free 3D path planning   & Critical obstacles and surrounding point set \\
\hline
\cite{guo2021three} & 3D obstacle avoidance for UAV & Circular Arc Trajectory Geometric Avoidance (CTGA) algorithm \\
\hline
\cite{ourari2022nearest} &  Decentralized collision avoidance  &  Based on starling flocks' behavior \\
\hline
\cite{huang2023e2copre} &  Energy efficient collision avoidance  & Artificial potential field and particle swarm planning \\
\hline
\end{tabular}
\end{table}

\begin{table}[htbp]
\centering
\caption{Summary on UAV Batching.}
\vspace{2mm}
\fontsize{9pt}{12pt}\selectfont
\label{tab:uav_batching}
\begin{tabular}{p{2.4cm} p{4.5cm} p{4.5cm} }
\hline
Studies  & Focus & Key Techniques/Methods Employed \\
\hline
\cite{fabra2020efficient} & Swarm takeoff procedure for VTOL of UAVs & Heuristic for optimizing UAV positions and implementing a sequential phased takeoff \\
\hline
\cite{hernandez2021kuhn} &  Swarm takeoff optimization &  Kuhn-Munkres algorithm for optimal, low-computation assignments  \\
\hline
\cite{sastre2022collision} &  Trajectory-based collision detection algorithm and batch generation mechanism & Grouping of UAVs into batches to enable simultaneous take-offs \\
\hline
\cite{sastre2022safe} &  Improved collision detection algorithm  & Introduction of novel methods: CSTH and ED-CSTH\\
\hline
\end{tabular}
\end{table}

{\it Third} and finally, past work done on batching is given in Table~\ref{tab:uav_batching}. The table shows that all papers employ very similar techniques. However, \\
\cite{hernandez2021kuhn} and \cite{sastre2022collision} are more popular. We directly adopt this approach and integrate it with the collision avoidance approach discussed above.

\section{Methodology} \label{Methodology}

\begin{figure}[htbp]
    \centering 
    \includegraphics[ height=2.5in]{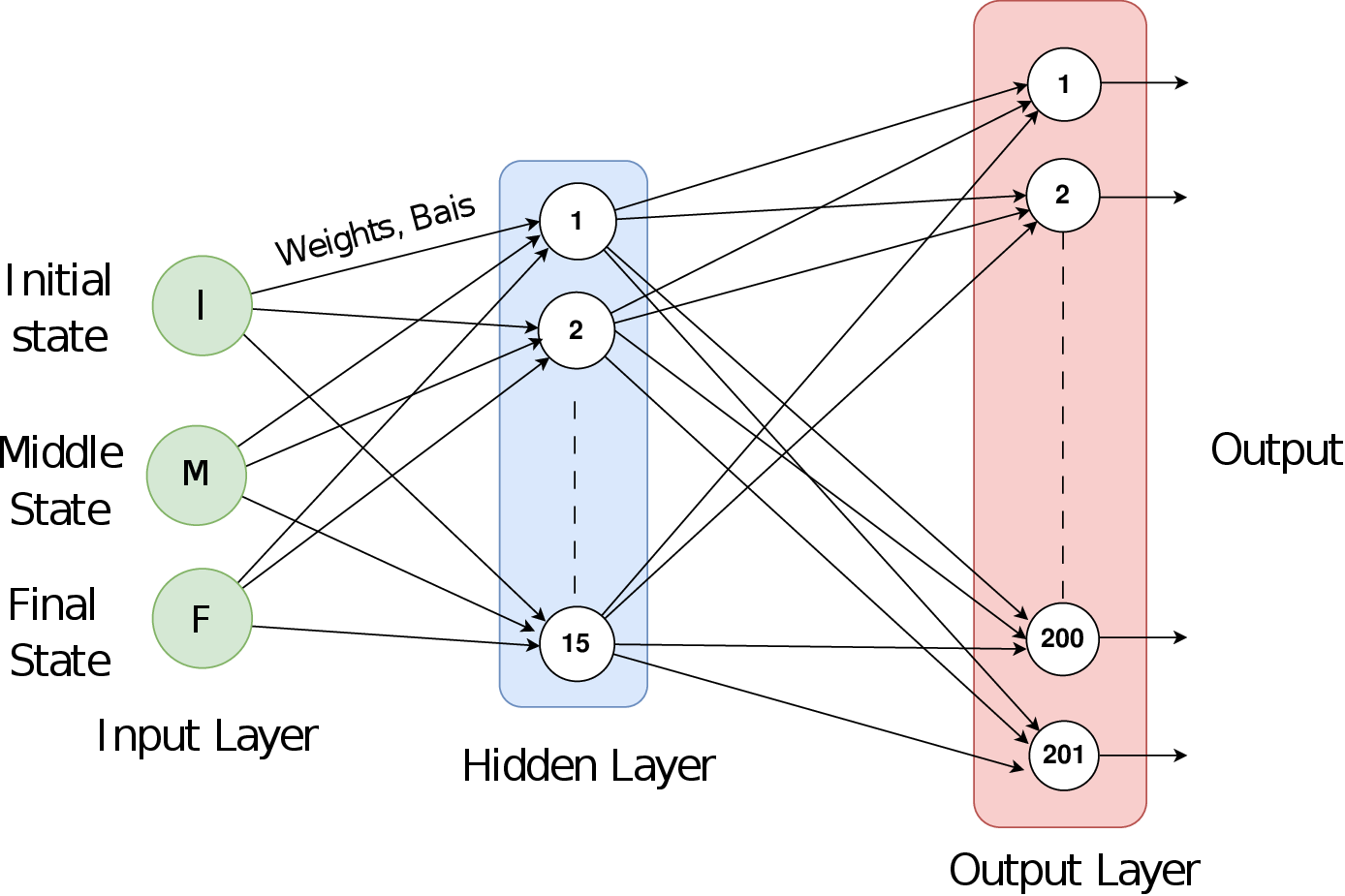}
    \caption{FFNN Architecture. }
    \label{fig:FFNN}
\end{figure}

Our research employs a FFNN architecture, as illustrated in Figure~\ref{fig:FFNN}, comprising an input layer, a hidden layer, and an output layer. The input layer of the FFNN is structured to incorporate the initial, intermediate, and final states of the UAV across all three spatial coordinates: X, Y, and Z. These states collectively form a multidimensional input vector that encapsulates the complete trajectory information of the UAV. 

% \subsection{Hidden Layer}
The hidden layer plays a pivotal role in capturing and representing complex patterns within the input data. Comprising 15 neurons, this layer employs various activation functions to nonlinearly transform the input, facilitating the extraction of relevant features and patterns essential for accurate prediction. This is the layer where we experiment with different types of activation functions and propose a new activation function as well, which is discussed below.

Responsible for producing the final predictions, the output layer is meticulously designed with 201 output neurons which comprise of linear polynomials. The deliberate choice of linear activation functions in this layer aligns with the nature of regression tasks, where the objective is to predict continuous numerical values. We don't change these activation functions in this layer.

To effectively train our neural network architecture, we employ the Levenberg-Marquardt backpropagation algorithm. This optimization technique combines the advantages of both the gradient descent and Gauss-Newton methods, making it well-suited for training neural networks, especially in cases where convergence and speed are of paramount importance as here.

The combination of a diverse set of activation functions in the hidden layer and the linear output layer forms a comprehensive framework for our research, allowing us to address the specific complexities and intricacies of our dataset. The rest of this section has three parts. In Section \ref{Activation Function}, we describe an array of activation functions, including the newest ones. In Section \ref{Novel_Activation_function}, we describe our novel AdaptoSwelliGauss activation function, in Section \ref{icdab}, we discuss our ICDAB technique.

\subsection{Activation Functions} \label{Activation Function}

In a neural network, an activation function plays a crucial role by applying a mathematical operation to the weighted sum of inputs. This introduces non-linearity, which is essential for the network to comprehend intricate patterns and connections within the data. Essentially, the activation function decides whether a neuron should fire or remain inactive, thereby impacting the flow of information throughout the network. In this sub-section, we delve into the various standard activation functions commonly employed in neural networks.

Here, \( x \) represents the input to each 
 activation function.

\subsubsection{Sigmoid}
The Sigmoid function maps input values into a probability range between 0 and 1, which is ideal for binary classification outputs. The formula for this is
\begin{equation}
    f(x) = \frac{1}{1 + e^{-x}}.
\end{equation}
Its major drawback is the vanishing gradient problem.
\subsubsection{Tanh}
Tanh extends the Sigmoid shape, mapping inputs to a range between -1 and 1, which is zero-centered. Thus, it typically results in faster convergence during training and is chosen for layers that benefit from data normalization. The formula for this is
\begin{equation}
    f(x) = \frac{e^x - e^{-x}}{e^x + e^{-x}}.
\end{equation}
Like Sigmoid, it suffers from vanishing gradients, affecting its utility in deep networks.

\subsubsection{ReLU}
ReLU addresses some of the critical issues of earlier activation functions by enabling faster training and deeper network compatibility thanks to its simple computation. It prevents the vanishing gradient issue, promoting healthy gradient levels during learning. It is given by
\begin{equation}
    f(x) = \max(0, x).
\end{equation}
 Nonetheless, ReLU is susceptible to the ``dying ReLU" problem, where neurons can irreversibly cease activity if they stop firing across many data points.

\subsubsection{Leaky ReLU}
Leaky ReLU modifies ReLU to allow a small input gradient when the unit is inactive. This modification ensures that all neurons have the opportunity to update during training, thereby avoiding the dying ReLU problem. It is given by
\begin{equation}
    f(x) = \max(\alpha x, x).
\end{equation}
where \(\alpha\) is a small coefficient. This function requires careful tuning of \(\alpha\), adding complexity to the network's training process.

\subsubsection{Swish}
Swish is a newer function that blends input and Sigmoid output. Its smooth gradient helps avoid issues seen in ReLU variants, maintaining robustness during deep network training. It is given by
\begin{equation}
    f(x) = x \cdot \sigma(\beta x),
\end{equation}
where $\sigma$ is the Sigmoid function and \(\beta\) is a trainable parameter. 

\subsubsection{Maxout Activation}
Maxout, proposed as a generalization of ReLU and its variants, is highly flexible, allowing the model to learn a piecewise linear approximation of any function. It is given by
\begin{equation}
    f(x) = \max(w_1 \cdot x + b_1, w_2 \cdot x + b_2,\ldots  ).
\end{equation}

\subsubsection{Elliot Activation}
Elliot activation quickly responds to changes in input values, making it particularly useful for tasks involving crucial variations. It is given by
\begin{equation}
    f(x) = \frac{x}{1 + |x|}.
\end{equation}

% Elliot's design helps to balance responsiveness with computational efficiency, which is beneficial for real-time processing.

\subsection{Novel Activation Function} \label{Novel_Activation_function}
We propose the activation function as below.
% The true strength of AdaptoSwelliGauss lies in the seamless orchestration of its components. Swish and Elliott work synergistically to adapt to both gradual and sudden changes in trajectory behavior, ensuring a comprehensive representation of UAV movements. The \(scale\) and \(shift\)  Gaussian component provides stability and continuity by smoothing activation transitions even in the presence of noise. The dynamic nature of AdaptoSwelliGauss not only allows it to capture trajectory nuances but also to respond intelligently to the varying challenges posed by real-world UAV movements.

% Here's a concise breakdown of its components and their collective contribution:

%\textbf{Activation Function Formula}:
%The AdaptoSwelliGauss activation function is formulated as follows:

\begin{equation}
\text{AdaptoSwelliGauss}(x)=
\begin{cases}
\text{Swish}(x), & \text{if } x \leq \alpha \\
\text{Elliott}(x) \times \text{Scale.Shift.Gaussian}(x), & \text{otherwise}
\end{cases}
\end{equation}

In this formulation,
\begin{itemize}
  \item $x$ represents the input to the activation function,
  \item $\text{Swish}(x)$ is the output when the Swish activation is applied to input $x$,
  \item $\text{Elliott}(x)$ is the output when the Elliott activation is applied to input $x$,
  \item $\text{Scale.Shift.Gaussian}(x)$ is the output of the scaled and shifted Gaussian function applied to input \(x\).
  \item $\alpha$ is the hyperparameter adjusted based on experimentation and learning.
\end{itemize}

\begin{figure}[htbp]
    \centering
    \includegraphics[height=2.5in]{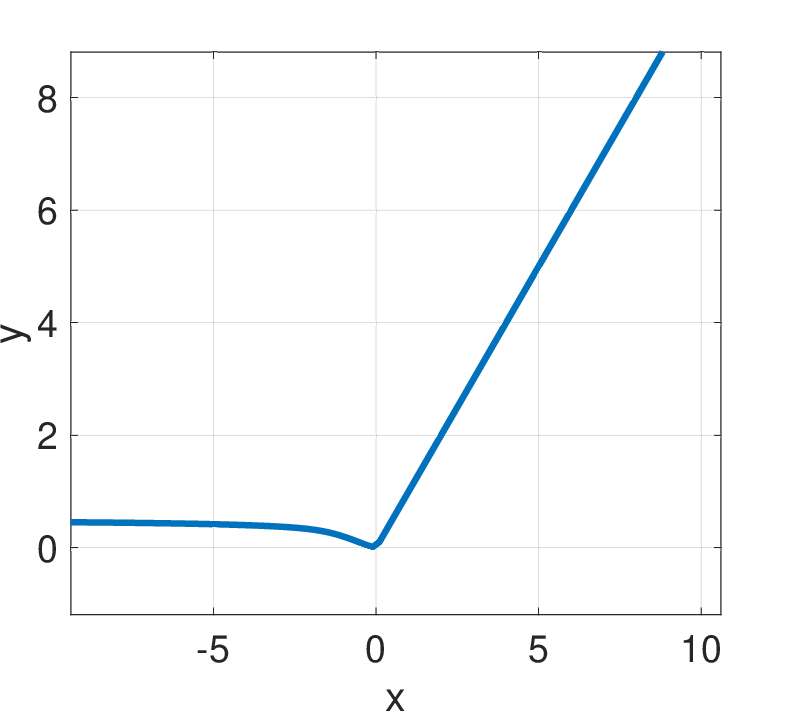}
    \caption{AdaptoSwelliGauss activation function.}
    \label{fig:adapto}
\end{figure}

The benefit of using the different components described above is discussed next.

% AdaptoSwelliGauss in Figure~\ref{fig:adapto} akin to ReLU and Swish, is unbounded above and bounded below, accommodating a diverse range of positive values. It inherits smoothness and non-monotonicity from Swish, facilitating differentiability and the capture of intricate relationships. When the input is less than equal to $\alpha$, it behaves like Swish, ensuring smooth activation. If the input surpasses $\alpha$, it combines Elliott and SSGaussian and remains inactive at zero input. In other cases, neurons consistently activate, allowing adaptability to input magnitudes. 
% This dual-behavior design contributes to versatility and responsiveness in capturing complex patterns within neural networks. The incorporation of Elliott and SSGaussian enhances the function's richness, making AdaptoSwelliGauss a valuable choice for diverse neural network architectures.

% \textbf{Why Swish, Elliott, and Gaussian Activation Function?}

\begin{enumerate}
  \item \textbf{Swish Activation}: As UAVs navigate through various trajectory phases, the Swish component seamlessly adjusts its activation levels to accommodate both gradual and subtle changes. This inherent adaptability ensures the activation function captures smooth transitions and maintains trajectory continuity, crucial for precise trajectory prediction.

  \item \textbf{Elliott Activation}: Complementing Swish's smoothness, the Elliott component infuses AdaptoSwelliGauss with heightened sensitivity to abrupt trajectory changes. When UAVs encounter sudden shifts in direction, altitude, or velocity, the Elliott activation swiftly responds with amplified gradients. Thereby delivering pinpoint accuracy during critical trajectory transitions.

  \item \textbf{Scaled and Shifted Gaussian}: The scaled and shifted Gaussian component adds a unique dimension to AdaptoSwelliGauss by mitigating the impact of noisy data points and preventing erratic activations that could compromise the accuracy of trajectory predictions. This 
 is implemented by adapting its parameters that help to fine-tune the activation function's behavior according to the noise characteristics present in the trajectory data.

\item \textbf{Dynamic Adaptation and Learning}:
The dynamic nature of AdaptoSwelliGauss is a critical asset. By adjusting the hyperparameter \(\alpha\)\ during training, the activation function fine-tunes its response based on the specifics of the trajectory data.
\end{enumerate}

AdaptoSwelliGauss, as in Figure~\ref{fig:adapto}, is unbounded above and bounded below. It is smooth, facilitating differentiability, and non-monotonic. 

\subsection{Integrated Collision Detection, Avoidance, and Batching (ICDAB)} \label{icdab}
As discussed in the introduction and literature review, this section presents our system designed to not only streamline the detection and avoidance of collisions among UAVs but to also optimize the deployment of UAV fleets. The novelty of this work lies in the integration of the below components. Each component has been designed to function synergistically, enhancing the overall efficacy of the UAV management system.

\begin{enumerate}
    \item \textbf{Collision Detection:} A geometric approach, as described in the works of \cite{elmkaiel2019collision}, \cite{lin2017sampling}, and \cite{huang2023e2copre}, is employed for collision detection. This method is widely accepted as standard.
    
    \item \textbf{Collision Avoidance:} Here, we build upon the framework established in \cite{guo2021three}. To facilitate integration with the subsequent component, a novel \texttt{tracking array} is introduced, which monitors the number of trajectory manipulations for each UAV. We also introduce a \texttt{batching list}, which stores the UAVs for whom this collision avoidance strategy fails.
    
    \item \textbf{Batching:} The batching technique, as outlined in \cite{sastre2022collision}, is adopted here. This approach is also recognized as a common standard in the literature, including in works like \cite{sastre2022safe} and \cite{10049400}. We synchronize this with the above collision avoidance component.
\end{enumerate}

Next, we discuss the above three components in detail.

\subsubsection{Collision Detection}

\begin{figure}[htbp]
    \centering
    \includegraphics[height=1.5in]{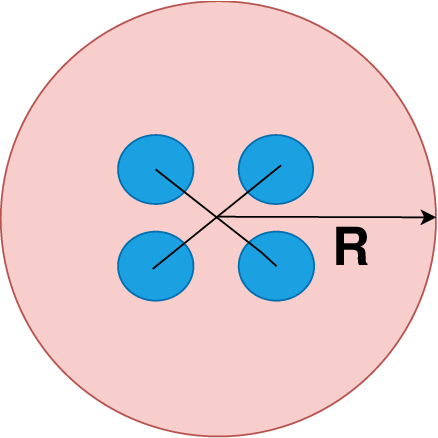}
    \caption{Collision sphere around the UAV.}
    \label{fig:collision_sphere}
\end{figure}

As practically common, we assume that all UAVs take off and fly simultaneously, following predefined trajectories between their respective pickup and delivery points. In our collision detection methodology, we employ a three-dimensional virtual sphere around each UAV referred to as the ``collision sphere''. This sphere is defined by a parameter ``R'' that denotes its radius (See Figure~\ref{fig:collision_sphere}). On a broader level, a collision is defined when any UAV collision sphere intersects with the collision sphere of another UAV. Next, we make this statement precise.

Let we be given a list of UAVs, their respective trajectories in terms of X, Y, and Z coordinates, their timestamps, and predefined point on the trajectory of each UAV where collision is to be checked (also called waypoints). Without loss of generality (WLOG) for a UAV pair, the standard algorithm calculates the Euclidean distance between their position at each waypoint along their paths and checks if this distance falls below a user-defined distance threshold when the timestamps of the UAV pair also meet a user-defined time threshold. If a collision is detected, then instead of the standard procedure of building a list of colliding pairs of UAVs, we apply the collision avoidance strategy discussed below. Finally, this procedure is repeated for all pairs.

%Each UAV, as depicted in Figure~\ref{fig:collision_sphere}, possesses its own distinct collision sphere. A collision is identified when any UAV within the airspace breaches the collision sphere of another UAV. Given a list of UAVs and their respective trajectories in terms of X, Y, and Z coordinates and timestamps, our algorithm systematically compares the trajectories of all UAV pairs. For each pair, it calculates the Euclidean distance between their positions at each waypoint along their paths and checks if this distance falls below a user-defined distance threshold when the timestamps of the UAV pair also meet a user-defined time threshold. If a collision is detected at any waypoint, the two UAVs are added to a \texttt{collision list}. The resulting \texttt{collision list} is then pruned to remove duplicate entries. We will next discuss the collision avoidance system.

\subsubsection{Collision Avoidance}
The primary objective here is to dynamically adjust the trajectory of two UAVs on a collision course. WLOG, we assume UAV1 and UAV2 are two UAVs that would collide. WLOG again, we change the trajectory of UAV1. This standard algorithm takes the following inputs: precise collision point of the two UAVs, and an array for padding adjustments to UAV1.

% This algorithm ensures collision prevention while concurrently monitoring and minimizing the number of trajectory manipulations. As shown in Figure~\ref{fig:collision_avoidance}, UAV1 and UAV2 are in the \texttt{collision list}. Without loss of generality, we change the trajectory of UAV1. The algorithm takes its precise collision point, an array for padding adjustments, and a \texttt{tracking array} to monitor the number of trajectory adjustments to it. 
% In the original paper \cite{guo2021three}, the padding array was not described in detail, which we pick up in a novel way, and \texttt{tracking array}, which we have proposed, is also new.

\begin{figure}[htbp]
    \centering
    \includegraphics[width=5in]{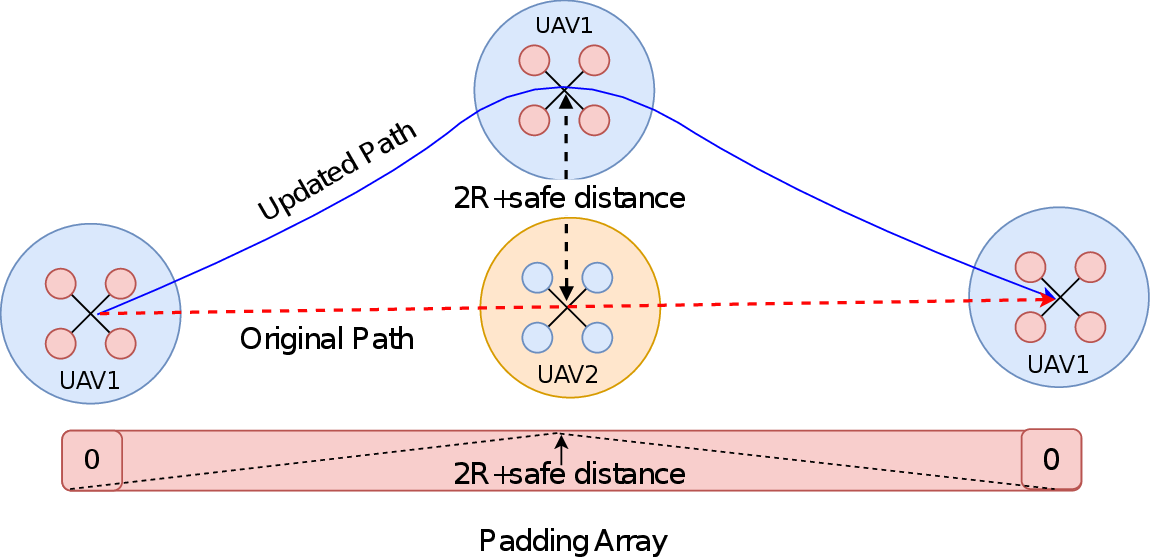}
    \caption{Collision avoidance between two UAVs.}
    \label{fig:collision_avoidance}
\end{figure}

 The distribution of values within the \texttt{padding array} follows a uniform pattern, incrementing uniformly from 0 to \texttt{2R+safe distance} and then uniformly decrementing back to 0. Here \texttt{safe distance} is the extra distance between the ``collision sphere'' of two UAVs. Next, UAV1 is checked for collision with all the remaining UAVs, and if there is one, then again collision avoidance strategy as above is applied. This strategic design, as shown in Figure~\ref{fig:collision_avoidance}, facilitates a gradual and balanced transition in UAV trajectories, optimizing collision avoidance effectiveness while maintaining trajectory simplicity.

 Next, we discuss our contribution that involves integration with batching. The trajectory manipulations, stored in an introduced \texttt{tracking array} for a UAV, are done up to a limit so as to avoid infinite adjustment (as happening in the standard approach). If this limit is exceeded for a UAV, then we look at the trajectory manipulation of the other UAV in the colliding pair. If adjustment for the other UAV also exceeds this limit, then both are moved to a \texttt{batching list}, as separate entries (not as a pair).

\subsubsection{Batching Mechanism}

% The traditional collision detection algorithm, as used in the batching paper, generates a list of possible collisions. However, since we are integrating collision detection and avoidance strategies from \cite{sastre2022safe}, we utilize the \texttt{batching list}, which is generated as discussed in the previous sections.

% The collision detection algorithm generates a list of possible collisions. We can now use this \texttt{batching list} to create batches of UAVs, which do not collide so that they can take off simultaneously. 
The batch generation algorithm is responsible for organizing UAVs into batches to facilitate coordinated flight. Its goal is to form collision-free batches that can be managed as a single unit. In the standard approach, the list of colliding pairs of UAVs is used as a input here.

Next, we describe our approach, which integrates with the above collision avoidance strategy. Here, the input to the batching algorithm is the \texttt{batching list}. The UAVs that are not on the \texttt{batching list} are free from any collision and outputted as one batch. 

Next, the $1^{st}$ UAV from the batching list is picked for the next batch. This $1^{st}$ UAV is checked for a collision with all the subsequent UAVs in the list ($2^{nd}$ UAV onwards).

WLOG, let the $i^{th}$ UAV does not collide with the $1^{st}$ UAV, then $1^{st}$ and $i^{th}$ UAVs are added to this next batch, and both are further checked for collision from $(i+1)^{th}$ UAV in the batching list. WLOG let $j^{th}$ UAV does not collide with both the $1^{st}$ and $i^{th}$ UAV. Then, $j^{th}$
UAV is also added to the next batch. Further, all three are checked for collision from $(j+1)^{th}$ UAV in the batching list.

This process is recursively done to determine this final next batch as well as all subsequent batches.

\section{Results} \label{Result}

Data for 500 UAVs is generated using a UAV simulator implemented with the UAV toolbox of Simulink in MATLAB. The simulation is run on a PC equipped with an Intel Core i7 12th generation processor, a 64-bit operating system, and 8GB of RAM. 

Each UAV’s trajectory is generated with a total of 201 waypoints, providing a rich and detailed dataset conducive to robust algorithm development. The configuration of initial and destination coordinates for each UAV is taken from \cite{lai2020machine}, which provides guidance on setting realistic ranges. The X and Y coordinates for the initial position range from [10-50], with a fixed initial Z-coordinate at 0. Destination coordinates are constrained within X, Y [200-300] with the Z-coordinate again fixed at zero. The maximum altitude achievable by the UAVs is set between [30-40]. 

\begin{figure}[htbp]
    \centering
    \includegraphics[width=5in]{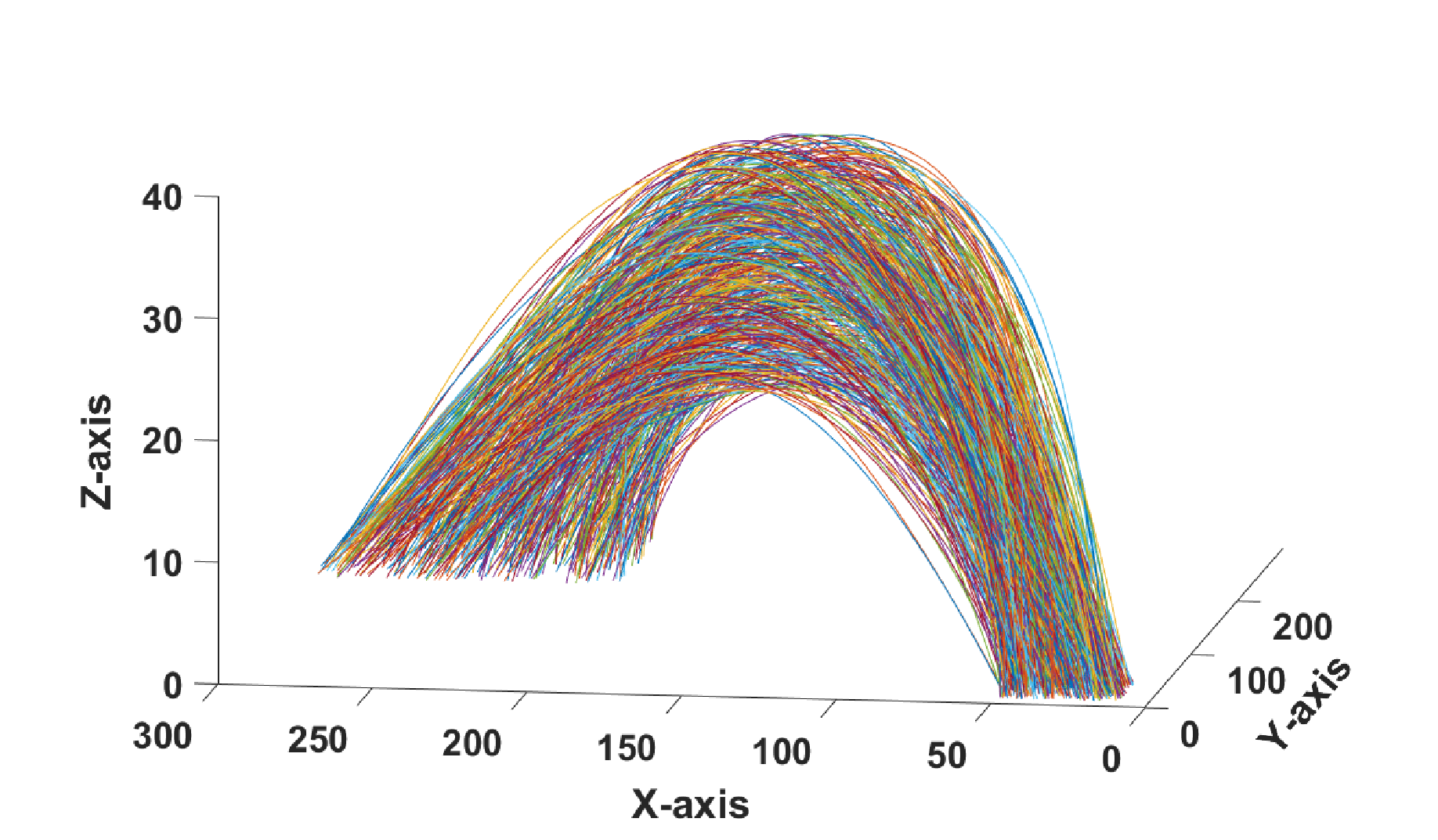}
    \caption{Dataset of 500 UAVs.}
    \label{fig:uav_dataset}
\end{figure}

A pictorial representation of the paths of all the UAVs is shown in Figure~\ref{fig:uav_dataset}. The dataset is partitioned into three distinct segments: 70\% is allocated for training, 15\% for validation, and the remaining 15\% for testing. The computational process iterates over 1000 epochs to optimize outcomes.

The choice of a loss function is pivotal in training neural networks as it quantifies the disparity between predicted and actual values. An effectively chosen loss function facilitates efficient learning and generalization, mitigating the risk of overfitting and enhancing model robustness. Loss functions are task-specific; for instance, cross-entropy is typically used for classification tasks, while Mean Squared Error (MSE) is often employed for regression tasks.

To measure the quality of trajectory prediction, various types of loss functions given in Table \ref{summary_of _loss_function}. Here, \( y_i \) represents the actual value, \( \hat{y}_i \) denotes the predicted value, and \( n \) symbolizes the number of observations. Also, the first two measures have been used extensively in the literature and the remaining three are less explored in this context.

\begin{table}[htbp]
\centering
\caption{Summary of Loss Functions.}
\vspace{2mm}
\label{summary_of _loss_function}
\fontsize{9pt}{12pt}\selectfont
\label{tab:collision_avoidance}
\begin{tabular}{p{2.5cm} p{2cm} p{6.5cm}}
\hline
S. No.  & Loss Function & \hspace{25mm}Formula \\
\hline
1. & Mean Squared Error (MSE) & \begin{equation*}
    MSE = \frac{1}{n} \sum_{i=1}^{n} (y_i - \hat{y}_i)^2.
  \end{equation*} \\
\hline
2. &  Root MSE (RMSE) &  \begin{equation*}
    RMSE = \sqrt{MSE}.
  \end{equation*} \\
\hline
3. &  Mean Absolute Error (MAE)   & \begin{equation*}
    MAE = \frac{1}{n} \sum_{i=1}^{n} |y_i - \hat{y}_i|.
  \end{equation*}\\
\hline
4. & Mean Absolute Percentage Error (MAPE) & \begin{equation*}
    MAPE = \frac{1}{n} \sum_{i=1}^{n} \left|\frac{y_i - \hat{y}_i}{y_i}\right| \times 100\%.
  \end{equation*} \\
\hline
5. & Symmetric MAPE (SMAPE) &  \begin{equation*}
    SMAPE = \frac{1}{n} \sum_{i=1}^{n} \frac{2|y_i - \hat{y}_i|}{|y_i| + |\hat{y}_i|} \times 100\%.
  \end{equation*} \\
\hline
\end{tabular}
\end{table}

Although MSE is the most commonly used loss function in the literature, according to  \cite{kreinovich2014estimate} and \cite{chicco2021coefficient}, SMAPE provides a symmetric measure that offers a better comparison in
certain scenarios. We use SMAPE in conjunction with MSE to understand
the performance of the dataset comprehensively.

The rest of this section has two parts. In Section \ref{section_4.1}, we present appropriate loss function values on the standard activation functions and our novel AdaptoSwelliGauss activation function (from Sections \ref{Activation Function} and \ref{Novel_Activation_function}). In Section \ref{section_4.2}, we give experimental results using ICDAB techniques (from Section \ref{icdab}) 

\subsection{Relevant Loss Functions Values Using Different Activations } \label{section_4.1}

For these experiments, the values of hyperparameters $\beta$, $\alpha$, \(scale\), and \(shift\) that work best for us are the median of the input to the activation functions, and are given by $0.5, 0.14, 0.5,$ and $0.25$, respectively.

X-coordinates results are given in Table~\ref{tab:error_metrics_x}. Among the standard activation functions, the ReLU and Elliot activation functions demonstrate the best performance achieving MSEs of \(1.733 \times 10^{-10}\) and \(3.605 \times 10^{-11}\) respectively, and SMAPEs of 0.19\% and 0.15\%, respectively. AdaptoSwelliGauss substantially surpasses both, attaining an MSE of \(3.059 \times 10^{-14}\) and a SMAPE of 0.14\%.

\renewcommand{\arraystretch}{1.5}
\begin{table}[htbp]
\caption{Error Metrics for the X-coordinate.}
\vspace{2mm}
\centering
\begin{tabular}{ c l c c }
\hline
\textbf{No.} & \textbf{Activation Function} & \textbf{MSE} & \textbf{SMAPE} \\ \hline
1            & Sigmoid                     & $9.200 \times 10^{-5}$ & 0.16\%  \\ \hline
2            & Tanh                        & $1.406 \times 10^{-7}$ & 0.29\%  \\ \hline
3            & ReLU                        & $1.733 \times 10^{-10}$ & 0.19\%  \\ \hline
4            & Leaky ReLU                  & $5.510 \times 10^{-5}$ & 0.21\% \\ \hline
5            & Swish                       & $1.638 \times 10^{-9}$ & 0.20\%  \\ \hline
6            & Maxout                      & $3.462 \times 10^{-6}$ & 0.15\%  \\ \hline
7            & Elliot                      & $3.605 \times 10^{-11}$ & 0.15\%  \\ \hline
\rowcolor[HTML]{32CB00} 
8            & AdaptoSwelliGauss           & $3.059 \times 10^{-14}$ & 0.14\%  \\ \hline
\end{tabular}
\label{tab:error_metrics_x}
\end{table}

% \begin{figure}[htbp]
%     \centering
%     \includegraphics[width=5in]{Figures/Results/X_best.jpg}
%     \caption{MSE on X-coordinates for train, test and validation}
%     \label{fig:MSE_X}
% \end{figure}

% Figure~\ref{fig:MSE_X} captures the training, validation, and testing MSE errors at each epoch, showcasing its superior performance.
For Y-coordinates, are given in Table~\ref{tab:error_metrics_y}. Among the standard activation functions, Swish and Elliot activation functions exhibit very good performance with MSE values of \(6.805 \times 10^{-10}\) and \(3.780 \times 10^{-9}\) respectively, and SMAPE values of 0.21\% and 0.18\%, respectively. AdaptoSwelliGauss outperforms them, achieving an MSE of \(5.127 \times 10^{-11}\) and a SMAPE of 0.15\%. 

% Please add the following required packages to your document preamble:
% \usepackage{multirow}
\renewcommand{\arraystretch}{1.5}
\begin{table}[htbp]
\caption{Error Metrics the for Y-coordinate.}
\vspace{2mm}
\centering
\begin{tabular}{ c l c c }
\hline
\textbf{No.} & \textbf{Activation Function} & \textbf{MSE} & \textbf{SMAPE} \\ \hline
1            & Sigmoid                     & $1.822 \times 10^{-8}$ & 0.17\%  \\ \hline
2            & Tanh                        & $8.664 \times 10^{-7}$ & 0.68\%  \\ \hline
3            & ReLU                        & $1.793 \times 10^{-8}$ & 0.29\%  \\ \hline
4            & Leaky ReLU                  & $2.934 \times 10^{-6}$ & 0.29\% \\ \hline
5            & Swish                       & $6.805 \times 10^{-10}$ & 0.21\%  \\ \hline
6            & Maxout                      & $2.192 \times 10^{-6}$ & 0.18\%  \\ \hline
7            & Elliot                      & $3.780 \times 10^{-9}$ & 0.18\%  \\ \hline
\rowcolor[HTML]{32CB00} 
8            & AdaptoSwelliGauss           & $5.127 \times 10^{-11}$ & 0.15\%  \\ \hline
\end{tabular}
\label{tab:error_metrics_y}
\end{table}

% \begin{figure}[htbp]
%     \centering
%     \includegraphics[width=5in]{Figures/Results/y.jpg}
%     \caption{MSE on Y-coordinates for train, test and validation}
%     \label{fig:MSE_Y}
% \end{figure}

% Figure~\ref{fig:MSE_Y} captures the training, validation, and testing MSE errors at each epoch, showcasing its optimal performance.
For Z-coordinates, results are given in Table~\ref{tab:error_metrics_z}.  Among the standard activation functions Swish and Maxout activation functions stand out with superior performance metrics achieving MSEs of \(2.182 \times 10^{-13}\) and \(2.630 \times 10^{-12}\), respectively, and SMAPE values of 0.13\% for both. AdaptoSwelliGauss slightly surpasses these, attaining an MSE of \(1.739 \times 10^{-13}\) and a SMAPE of 0.12\%.

\begin{table}[htbp]
\caption{Error Metrics for the Z-coordinate. }
\vspace{2mm}
\centering
\begin{tabular}{c l c c }
\hline
\textbf{No.} & \textbf{Activation Function} & \textbf{MSE} & \textbf{SMAPE} \\ \hline
1            & Sigmoid                     & $4.588 \times 10^{-11}$ & 0.34\%  \\ \hline
2            & Tanh                        & $8.031 \times 10^{-7}$ & 0.61\%  \\ \hline
3            & ReLU                        & $5.851 \times 10^{-9}$ & 0.31\%  \\ \hline
4            & Leaky ReLU                  & $1.344 \times 10^{-7}$ & 0.29\% \\ \hline
5            & Swish                       & $2.182 \times 10^{-13}$ & 0.13\%  \\ \hline
6            & Maxout                      & $2.630 \times 10^{-12}$ & 0.13\%  \\ \hline
7            & Elliot                      & $4.851 \times 10^{-8}$ & 0.36\%  \\ \hline
\rowcolor[HTML]{32CB00} 
8            & AdaptoSwelliGauss           & $1.739 \times 10^{-13}$ & 0.12\%  \\ \hline
\end{tabular}
\label{tab:error_metrics_z}
\end{table}

% \begin{figure}[htbp]
%     \centering
%     \includegraphics[width=5in]{Figures/Results/Z_best.jpg}
%     \caption{MSE on Z-coordinates for train, test and validation}
%     \label{fig:MSE_Z}
% \end{figure}

% Figure~\ref{fig:MSE_Z} illustrates the training, validation, and testing phases, emphasizing its best performance.

Across all three coordinates, among the standard activation functions, ReLU, Swish, and Elliot give the best results. AdaptoSwelliGauss {\it consistently demonstrates substantially superior performance compared to all of these}, highlighting its robustness and effectiveness in enhancing the accuracy and predictive capabilities of the neural network across different dimensions.

\subsection{Results On ICDAB} \label{section_4.2}

When we apply collision detection as described in Section 3.3.1. we detect 262 collisions. Here the value of ``R'' is $0.5$. Next, we apply the original (or standard) collision avoidance as initially described in Section 3.3.2 (i.e., one where infinite trajectory manipulation is allowed), and the algorithm gets stuck in a loop. 

Thus, further, we apply our modified collision avoidance algorithm (as given at the end of Section 3.3.2) along with batching (from Section 3.3.3). This integration leads to substantial improvement in both components.

{\it First}, the application of the modified collision avoidance (i.e., up to ten trajectory manipulations allowed) leads to convergence of the algorithm, eventually resulting in only 41 possible collisions. Batching brings these collisions down to zero. Here, the safe radius taken is 0.5.

{\it Second}, results for batching integrated with the modified collision avoidance are given in Table~\ref{tab:ecda_results}. As evident from this table, the number of batches required is reduced from 7 to 5, while the maximum number of UAVs per batch increased from 38 to 315.

% In this section, we initially experimented with our collision avoidance algorithm without imposing limits on the number of trajectory manipulations. This approach led to over 184 UAVs experiencing more than 300 trajectory manipulations each, highlighting a significant challenge in terms of scalability and efficiency. Motivated by these results, we implemented constraints on the number of trajectory manipulations and integrated the collision detection with batching to optimize the process.

% We also explored the effectiveness of collision avoidance in reducing the necessity for extensive batching. We set the collision and safe radii at 0.5, which are appropriate for medium-sized UAVs commonly used in real-world applications. These radii are crucial for determining the proximity at which collision avoidance measures are triggered.

\begin{table}[htbp]
\caption{Result before and after applying ICDAB.}
\vspace{2mm}
\label{tab:ecda_results}
\centering
\begin{tabular}{p{4cm} p{5cm} p{3cm}}
\hline
\textbf{Metric} & \textbf{Before Collision Avoidance} & \textbf{After Collision Avoidance} \\ \hline
Number of Batches & 7 & 5 \\ \hline
Maximum number of UAVs per Batch & 38 & 315 \\ \hline
\end{tabular}
\end{table}

% As shown in Table~\ref{tab:ecda_results}, implementing collision avoidance with a limit on trajectory manipulations significantly improves the system's efficiency. The number of detected collisions decreased dramatically from 262 to 47. Additionally, the need for multiple batches was reduced from 7 to 5, while the maximum number of UAVs per batch increased substantially from 38 to 315. These enhancements demonstrate the advantages of integrating collision avoidance with batching, leading to a more efficient handling of UAVs in congested spaces.

Next, we show the importance of carefully calibrating the safe radius in our ICDAB algorithm. As shown in Table~\ref{tab:safe_radius}, increasing the safe radius initially results in a decrease in the number of colliding UAVs, decrease in number of batches, and increase in the maximum number of UAVs per batch, achieving optimal results at a safe radius of 2.3. At this radius, the number of colliding UAVs is reduced to 33, the number of batches is minimized to 4, and the maximum number of UAVs per batch increases to 396. However, beyond this optimal point, further increases in the safe radius lead to a reverse trend, where both the number of colliding UAVs and the number of batches begin to increase again, while the maximum number of UAVs per batch decreases. These results are subject to the characteristics and dynamics of the specific dataset under consideration, and different datasets may yield different optimal safe radius values. However, there is a scientific reason for this observed trend.
\begin{table}[htbp]
\caption{ICDAB Outcomes with Different Safe Radius Values.}
\vspace{2mm}
\centering
\begin{tabular}{p{1.4cm} p{2cm} p{2cm}  p{2cm}}
% \begin{tabular}{|l|l|l|l|}
\hline
Safe Radius & Number of Collision After Avoidance & Number of Batches & Max Number of UAVs per Batch \\ \hline
0.5         & 47             & 5                 & 315                            \\ \hline
0.7         & 46             & 6                 & 321                            \\ \hline
1.9         & 39             & 7                 & 326                            \\ \hline
2.1         & 35             & 4                 & 378                            \\ \hline 
\rowcolor[HTML]{32CB00} 
2.3         & 33             & 4                 & 396                            \\ \hline
2.5         & 42             & 6                 & 335                            \\ \hline
2.7         & 61             & 8                 & 305                            \\ \hline
2.9         & 65             & 10                & 278                            \\ \hline
\end{tabular}
\label{tab:safe_radius}
\end{table}

\begin{figure}[htbp]
    \centering
    \includegraphics[height=3.2in]{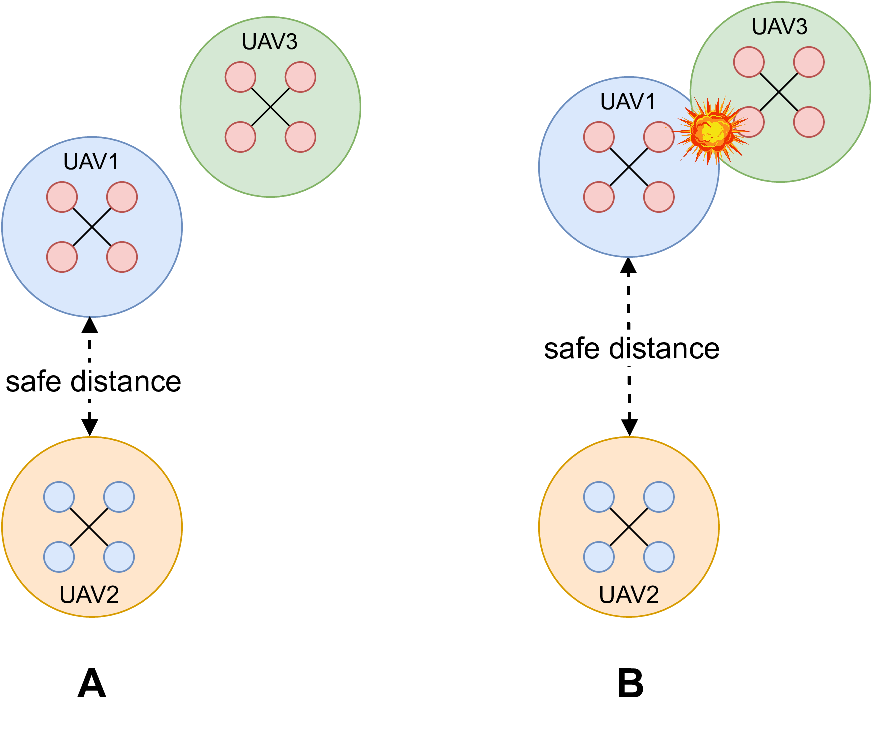}
    \caption{\textbf{A:} ICDAB with small safe radius, \textbf{B:} ICDAB with large safe radius.}
    \label{fig:ICDAB_safe_radius}
\end{figure}

This is because of unintended consequences of over-adjusting UAV trajectories at larger safe radius (see Figure~\ref{fig:ICDAB_safe_radius}). As in the given figure, when the trajectory of UAV1 is significantly altered to avoid UAV2, it may inadvertently collide with UAV3, which was previously not on a collision course with UAV1. This has a cascading effect on all the UAVs. 

\section{Conclusion} \label{Conclusion}
In this paper, we focus on the development of an intelligent framework for predicting and optimizing the trajectory of a fleet of UAVs. Traditional methods have limitations in accurately predicting paths as well as efficiently avoiding collisions for large fleets of UAVs. We address these challenges here.

 In our research, we {\it first} conduct a comprehensive comparative analysis by applying various activation functions to a FFNN with a single hidden layer.

{\it Second} is the introduction of a novel activation function called AdaptoSwelliGauss. This activation function substantially outperforms commonly used counterparts in terms of trajectory accuracy. 

{\it Third}, we improve existing methods for collision detection and avoidance. Detection is straightforward while avoidance is challenging. We can avoid collision between UAVs by either changing their trjaectories or sending them in batches. Each of which has its drawbacks. We uniquely combine those technique leading to improved performance.

Future work involves improving the underlying mathematical optimisation \cite{ahuja2008Mixed}; exploring Convolutional Neural Networks (CNNs) for enhanced trajectory accuracy \cite{li2021survey}; integrating advanced sensors like LiDAR for real-time collision detection \cite{wei2018lidar}; using approximate computing in neural networks \cite{gupta2020ApproxAccess, gupta2020ApproxProc}; leveraging multi-agent reinforcement learning for collaborative decision-making \cite{canese2021multi}; exploiting implicit relation between different drone trajectories \cite{kim2005Explain};  
% adapting algorithms for dynamic environments; 
and conducting real-world validation studies to ensure regulatory compliance and operational efficiency in autonomous UAV operations.

% \bibliographystyle{unsrt}
% \bibliography{References.bib}

% \vfill

% \end{document}

%% The Appendices part is started with the command \appendix;
%% appendix sections are then done as normal sections
%% \appendix

%% \section{}
%% \label{}

%% If you have bibdatabase file and want bibtex to generate the
%% bibitems, please use
%%
 \bibliographystyle{elsarticle-harv} 
 \bibliography{References}  

\begin{thebibliography}{29}
\expandafter\ifx\csname natexlab\endcsname\relax\def\natexlab#1{#1}\fi
\providecommand{\url}[1]{\texttt{#1}}
\providecommand{\href}[2]{#2}
\providecommand{\path}[1]{#1}
\providecommand{\DOIprefix}{doi:}
\providecommand{\ArXivprefix}{arXiv:}
\providecommand{\URLprefix}{URL: }
\providecommand{\Pubmedprefix}{pmid:}
\providecommand{\doi}[1]{\href{http://dx.doi.org/#1}{\path{#1}}}
\providecommand{\Pubmed}[1]{\href{pmid:#1}{\path{#1}}}
\providecommand{\bibinfo}[2]{#2}
\ifx\xfnm\relax \def\xfnm[#1]{\unskip,\space#1}\fi
%Type = Article
\bibitem[{Ahuja et~al.(2008)Ahuja, Watson and Billups}]{ahuja2008Mixed}
\bibinfo{author}{Ahuja, K.}, \bibinfo{author}{Watson, L.T.},
  \bibinfo{author}{Billups, S.C.}, \bibinfo{year}{2008}.
\newblock \bibinfo{title}{Probability-one homotopy maps for mixed
  complementarity problems}.
\newblock \bibinfo{journal}{Computational Optimization and Applications}
  \bibinfo{volume}{41}, \bibinfo{pages}{363 -- 375}.
%Type = Article
\bibitem[{Canese et~al.(2021)Canese, Cardarilli, Di~Nunzio, Fazzolari,
  Giardino, Re and Span{\`o}}]{canese2021multi}
\bibinfo{author}{Canese, L.}, \bibinfo{author}{Cardarilli, G.C.},
  \bibinfo{author}{Di~Nunzio, L.}, \bibinfo{author}{Fazzolari, R.},
  \bibinfo{author}{Giardino, D.}, \bibinfo{author}{Re, M.},
  \bibinfo{author}{Span{\`o}, S.}, \bibinfo{year}{2021}.
\newblock \bibinfo{title}{Multi-agent reinforcement learning: A review of
  challenges and applications}.
\newblock \bibinfo{journal}{Applied Sciences} \bibinfo{volume}{11},
  \bibinfo{pages}{4948}.
%Type = Article
\bibitem[{Chicco et~al.(2021)Chicco, Warrens and
  Jurman}]{chicco2021coefficient}
\bibinfo{author}{Chicco, D.}, \bibinfo{author}{Warrens, M.J.},
  \bibinfo{author}{Jurman, G.}, \bibinfo{year}{2021}.
\newblock \bibinfo{title}{The coefficient of determination {R}-squared is more
  informative than {SMAPE}, {MAE}, {MAPE}, {MSE} and {RMSE} in regression
  analysis evaluation}.
\newblock \bibinfo{journal}{{PeerJ} {C}omputer {S}cience}
  \bibinfo{volume}{7:e623}.
%Type = Inproceedings
\bibitem[{Elmkaiel and Serebrenny(2019)}]{elmkaiel2019collision}
\bibinfo{author}{Elmkaiel, G.}, \bibinfo{author}{Serebrenny, V.},
  \bibinfo{year}{2019}.
\newblock \bibinfo{title}{Collision avoidance algorithm for a quadcopters
  swarm}, in: \bibinfo{booktitle}{AIP Conference Proceedings},
  \bibinfo{organization}{AIP Publishing}. p. \bibinfo{pages}{190006}.
%Type = Inproceedings
\bibitem[{Fabra et~al.(2020)Fabra, Wubben, Calafate, Cano and
  Manzoni}]{fabra2020efficient}
\bibinfo{author}{Fabra, F.}, \bibinfo{author}{Wubben, J.},
  \bibinfo{author}{Calafate, C.T.}, \bibinfo{author}{Cano, J.C.},
  \bibinfo{author}{Manzoni, P.}, \bibinfo{year}{2020}.
\newblock \bibinfo{title}{Efficient and coordinated vertical takeoff of {UAV}
  swarms}, in: \bibinfo{booktitle}{2020 IEEE 91st Vehicular Technology
  Conference (VTC2020-Spring)}, \bibinfo{organization}{IEEE}. pp.
  \bibinfo{pages}{1--5}.
%Type = Article
\bibitem[{Guo et~al.(2021)Guo, Liang, Wang, Sang and Wu}]{guo2021three}
\bibinfo{author}{Guo, J.}, \bibinfo{author}{Liang, C.}, \bibinfo{author}{Wang,
  K.}, \bibinfo{author}{Sang, B.}, \bibinfo{author}{Wu, Y.},
  \bibinfo{year}{2021}.
\newblock \bibinfo{title}{Three-dimensional autonomous obstacle avoidance
  algorithm for {UAV} based on circular arc trajectory}.
\newblock \bibinfo{journal}{International journal of aerospace engineering}
  \bibinfo{volume}{8819618}.
%Type = Article
\bibitem[{Gupta et~al.(2020)Gupta, Ullah, Ahuja, Tiwari and
  Kumar}]{gupta2020ApproxAccess}
\bibinfo{author}{Gupta, S.}, \bibinfo{author}{Ullah, S.},
  \bibinfo{author}{Ahuja, K.}, \bibinfo{author}{Tiwari, A.},
  \bibinfo{author}{Kumar, A.}, \bibinfo{year}{2020}.
\newblock \bibinfo{title}{{ALigN}: A highly accurate adaptive layerwise
  {Log\_2\_Lead} quantization of p{re\-trained} neural networks}.
\newblock \bibinfo{journal}{IEEE Access} \bibinfo{volume}{8},
  \bibinfo{pages}{118899}.
%Type = Article
\bibitem[{Han(2019)}]{han2019efficient}
\bibinfo{author}{Han, J.}, \bibinfo{year}{2019}.
\newblock \bibinfo{title}{An efficient approach to {3D} path planning}.
\newblock \bibinfo{journal}{Information Sciences} \bibinfo{volume}{478},
  \bibinfo{pages}{318--330}.
%Type = Inproceedings
\bibitem[{Hern{\'a}ndez et~al.(2021)Hern{\'a}ndez, Cec{\'\i}lia, Calafate, Cano
  and Manzoni}]{hernandez2021kuhn}
\bibinfo{author}{Hern{\'a}ndez, D.}, \bibinfo{author}{Cec{\'\i}lia, J.M.},
  \bibinfo{author}{Calafate, C.T.}, \bibinfo{author}{Cano, J.C.},
  \bibinfo{author}{Manzoni, P.}, \bibinfo{year}{2021}.
\newblock \bibinfo{title}{The {K}uhn-{M}unkres algorithm for efficient vertical
  takeoff of {UAV} swarms}, in: \bibinfo{booktitle}{2021 IEEE 93rd Vehicular
  Technology Conference (VTC2021-Spring)}, \bibinfo{organization}{IEEE}. pp.
  \bibinfo{pages}{1--5}.
%Type = Article
\bibitem[{Huang et~al.(2023)Huang, Zhang and Huang}]{huang2023e2copre}
\bibinfo{author}{Huang, S.}, \bibinfo{author}{Zhang, H.},
  \bibinfo{author}{Huang, Z.}, \bibinfo{year}{2023}.
\newblock \bibinfo{title}{{E}2{C}o{P}re: Energy efficient and cooperative
  collision avoidance for {UAV} swarms with trajectory prediction}.
\newblock \bibinfo{journal}{arXiv preprint arXiv:2303.06510} .
%Type = Article
\bibitem[{Jeong et~al.(2021)Jeong, You and Seok}]{jeong2021hazardous}
\bibinfo{author}{Jeong, S.}, \bibinfo{author}{You, K.}, \bibinfo{author}{Seok,
  D.}, \bibinfo{year}{2021}.
\newblock \bibinfo{title}{Hazardous flight region prediction for a small {UAV}
  operated in an urban area using a deep neural network}.
\newblock \bibinfo{journal}{Aerospace Science and Technology}
  \bibinfo{volume}{118}, \bibinfo{pages}{107060}.
%Type = Article
\bibitem[{Jiang et~al.(2022)Jiang, Li, Zhou, Lo, Chen and
  Wen}]{jiang2022neural}
\bibinfo{author}{Jiang, B.}, \bibinfo{author}{Li, B.}, \bibinfo{author}{Zhou,
  W.}, \bibinfo{author}{Lo, L.Y.}, \bibinfo{author}{Chen, C.K.},
  \bibinfo{author}{Wen, C.Y.}, \bibinfo{year}{2022}.
\newblock \bibinfo{title}{Neural network based model predictive control for a
  quadrotor {UAV}}.
\newblock \bibinfo{journal}{Aerospace} \bibinfo{volume}{9},
  \bibinfo{pages}{460}.
%Type = Incollection
\bibitem[{Kim et~al.(2005)Kim, Murthy, Ahuja, Vasile and Fox}]{kim2005Explain}
\bibinfo{author}{Kim, S.}, \bibinfo{author}{Murthy, U.},
  \bibinfo{author}{Ahuja, K.}, \bibinfo{author}{Vasile, S.},
  \bibinfo{author}{Fox, E.A.}, \bibinfo{year}{2005}.
\newblock \bibinfo{title}{Effectiveness of implicit rating data on
  characterizing users in complex information systems}, in:
  \bibinfo{editor}{Rauber, A.}, \bibinfo{editor}{Christodoulakis, S.},
  \bibinfo{editor}{Tjoa, A.M.e.} (Eds.), \bibinfo{booktitle}{Research and
  Advanced Technology for Digital Libraries (ECDL 2005), Lecture Notes in
  Computer Science}. \bibinfo{publisher}{Springer}. volume
  \bibinfo{volume}{3652}, pp. \bibinfo{pages}{186 -- 194}.
%Type = Article
\bibitem[{Kreinovich et~al.(2014)Kreinovich, Nguyen and
  Ouncharoen}]{kreinovich2014estimate}
\bibinfo{author}{Kreinovich, V.}, \bibinfo{author}{Nguyen, H.T.},
  \bibinfo{author}{Ouncharoen, R.}, \bibinfo{year}{2014}.
\newblock \bibinfo{title}{How to estimate forecasting quality: A
  system-motivated derivation of symmetric mean absolute percentage error
  {SMAPE} and other similar characteristics}.
\newblock \bibinfo{journal}{Departmental Technical Reports (CS)}
  \bibinfo{volume}{865}, \bibinfo{pages}{1--12}.
%Type = Book
\bibitem[{Lai(2020)}]{lai2020machine}
\bibinfo{author}{Lai, R.}, \bibinfo{year}{2020}.
\newblock \bibinfo{title}{A machine learning approach to trajectory planning
  for UAV}.
\newblock \bibinfo{publisher}{M.S. Thesis, Rensselaer Polytechnic Institute}.
%Type = Article
\bibitem[{Li et~al.(2021)Li, Liu, Yang, Peng and Zhou}]{li2021survey}
\bibinfo{author}{Li, Z.}, \bibinfo{author}{Liu, F.}, \bibinfo{author}{Yang,
  W.}, \bibinfo{author}{Peng, S.}, \bibinfo{author}{Zhou, J.},
  \bibinfo{year}{2021}.
\newblock \bibinfo{title}{A survey of convolutional neural networks: analysis,
  applications, and prospects}.
\newblock \bibinfo{journal}{IEEE transactions on neural networks and learning
  systems} \bibinfo{volume}{33}, \bibinfo{pages}{6999--7019}.
%Type = Article
\bibitem[{Lin and Saripalli(2017)}]{lin2017sampling}
\bibinfo{author}{Lin, Y.}, \bibinfo{author}{Saripalli, S.},
  \bibinfo{year}{2017}.
\newblock \bibinfo{title}{Sampling-based path planning for {UAV} collision
  avoidance}.
\newblock \bibinfo{journal}{IEEE Transactions on Intelligent Transportation
  Systems} \bibinfo{volume}{18}, \bibinfo{pages}{3179--3192}.
%Type = Inproceedings
\bibitem[{Ourari et~al.(2022)Ourari, Cui, Elshamanhory and
  Koeppl}]{ourari2022nearest}
\bibinfo{author}{Ourari, R.}, \bibinfo{author}{Cui, K.},
  \bibinfo{author}{Elshamanhory, A.}, \bibinfo{author}{Koeppl, H.},
  \bibinfo{year}{2022}.
\newblock \bibinfo{title}{Nearest-neighbor-based collision avoidance for
  quadrotors via reinforcement learning}, in: \bibinfo{booktitle}{2022
  International Conference on Robotics and Automation (ICRA)},
  \bibinfo{organization}{IEEE}. pp. \bibinfo{pages}{293--300}.
%Type = Article
\bibitem[{Qiu and Duan(2020)}]{qiu2020multi}
\bibinfo{author}{Qiu, H.}, \bibinfo{author}{Duan, H.}, \bibinfo{year}{2020}.
\newblock \bibinfo{title}{A multi-objective pigeon-inspired optimization
  approach to {UAV} distributed flocking among obstacles}.
\newblock \bibinfo{journal}{Information Sciences} \bibinfo{volume}{509},
  \bibinfo{pages}{515--529}.
%Type = Inproceedings
\bibitem[{Sarkar et~al.(2020)Sarkar, Totaro and Kumar}]{sarkar2020intelligent}
\bibinfo{author}{Sarkar, S.}, \bibinfo{author}{Totaro, M.W.},
  \bibinfo{author}{Kumar, A.}, \bibinfo{year}{2020}.
\newblock \bibinfo{title}{An intelligent framework for prediction of a
  {UAV}’s flight time}, in: \bibinfo{booktitle}{2020 16th International
  Conference on Distributed Computing in Sensor Systems (DCOSS)},
  \bibinfo{organization}{IEEE}. pp. \bibinfo{pages}{328--332}.
%Type = Inproceedings
\bibitem[{Sastre et~al.(2022a)Sastre, Wubben, Calafate, Cano and
  Manzoni}]{sastre2022collision}
\bibinfo{author}{Sastre, C.}, \bibinfo{author}{Wubben, J.},
  \bibinfo{author}{Calafate, C.T.}, \bibinfo{author}{Cano, J.C.},
  \bibinfo{author}{Manzoni, P.}, \bibinfo{year}{2022}a.
\newblock \bibinfo{title}{Collision-free swarm take-off based on trajectory
  analysis and {UAV} grouping}, in: \bibinfo{booktitle}{2022 IEEE 23rd
  International Symposium on a World of Wireless, Mobile and Multimedia
  Networks (WoWMoM)}, \bibinfo{organization}{IEEE}. pp.
  \bibinfo{pages}{477--482}.
%Type = Article
\bibitem[{Sastre et~al.(2022b)Sastre, Wubben, Calafate, Cano and
  Manzoni}]{sastre2022safe}
\bibinfo{author}{Sastre, C.}, \bibinfo{author}{Wubben, J.},
  \bibinfo{author}{Calafate, C.T.}, \bibinfo{author}{Cano, J.C.},
  \bibinfo{author}{Manzoni, P.}, \bibinfo{year}{2022}b.
\newblock \bibinfo{title}{Safe and efficient take-off of {VTOL} {UAV} swarms}.
\newblock \bibinfo{journal}{Electronics} \bibinfo{volume}{11},
  \bibinfo{pages}{1128}.
%Type = Inproceedings
\bibitem[{Ullah et~al.(2020)Ullah, Gupta, Ahuja, Tiwari and
  Kumar}]{gupta2020ApproxProc}
\bibinfo{author}{Ullah, S.}, \bibinfo{author}{Gupta, S.},
  \bibinfo{author}{Ahuja, K.}, \bibinfo{author}{Tiwari, A.},
  \bibinfo{author}{Kumar, A.}, \bibinfo{year}{2020}.
\newblock \bibinfo{title}{{L2L}: A highly accurate {Log\_2\_Lead} quantization
  of p{re\-trained} neural networks}, in: \bibinfo{booktitle}{2020 Design,
  Automation \& Test in Europe Conference \& Exhibition (DATE)},
  \bibinfo{organization}{IEEE}. pp. \bibinfo{pages}{979--982}.
%Type = Article
\bibitem[{Wan et~al.(2019)Wan, Tang and Lao}]{wan2019distributed}
\bibinfo{author}{Wan, Y.}, \bibinfo{author}{Tang, J.}, \bibinfo{author}{Lao,
  S.}, \bibinfo{year}{2019}.
\newblock \bibinfo{title}{Distributed conflict-detection and resolution
  algorithm for {UAV} swarms based on consensus algorithm and strategy
  coordination}.
\newblock \bibinfo{journal}{IEEE Access} \bibinfo{volume}{7},
  \bibinfo{pages}{100552--100566}.
%Type = Article
\bibitem[{Wei et~al.(2018)Wei, Cagle, Reza, Ball and Gafford}]{wei2018lidar}
\bibinfo{author}{Wei, P.}, \bibinfo{author}{Cagle, L.}, \bibinfo{author}{Reza,
  T.}, \bibinfo{author}{Ball, J.}, \bibinfo{author}{Gafford, J.},
  \bibinfo{year}{2018}.
\newblock \bibinfo{title}{{LiDAR} and camera detection fusion in a real-time
  industrial multi-sensor collision avoidance system}.
\newblock \bibinfo{journal}{Electronics} \bibinfo{volume}{7},
  \bibinfo{pages}{84}.
%Type = Article
\bibitem[{Wubben et~al.(2023)Wubben, Hernández, Cecilia, Imberón, Calafate,
  Cano, Manzoni and Toh}]{10049400}
\bibinfo{author}{Wubben, J.}, \bibinfo{author}{Hernández, D.},
  \bibinfo{author}{Cecilia, J.M.}, \bibinfo{author}{Imberón, B.},
  \bibinfo{author}{Calafate, C.T.}, \bibinfo{author}{Cano, J.C.},
  \bibinfo{author}{Manzoni, P.}, \bibinfo{author}{Toh, C.K.},
  \bibinfo{year}{2023}.
\newblock \bibinfo{title}{Assignment and take-off approaches for large-scale
  autonomous {UAV} swarms}.
\newblock \bibinfo{journal}{IEEE Transactions on Intelligent Transportation
  Systems} \bibinfo{volume}{24}, \bibinfo{pages}{4836--4847}.
%Type = Inproceedings
\bibitem[{Xiao et~al.(2019)Xiao, Zhao, He and Yu}]{xiao2019trajectory}
\bibinfo{author}{Xiao, K.}, \bibinfo{author}{Zhao, J.}, \bibinfo{author}{He,
  Y.}, \bibinfo{author}{Yu, S.}, \bibinfo{year}{2019}.
\newblock \bibinfo{title}{Trajectory prediction of {UAV} in smart city using
  recurrent neural networks}, in: \bibinfo{booktitle}{ICC 2019-2019 IEEE
  International Conference on Communications}, \bibinfo{organization}{IEEE}.
  pp. \bibinfo{pages}{1--6}.
%Type = Article
\bibitem[{Xu et~al.(2024)Xu, Xie, Luo, Zhang and Zhang}]{xu2024multi}
\bibinfo{author}{Xu, X.}, \bibinfo{author}{Xie, C.}, \bibinfo{author}{Luo, Z.},
  \bibinfo{author}{Zhang, C.}, \bibinfo{author}{Zhang, T.},
  \bibinfo{year}{2024}.
\newblock \bibinfo{title}{A multi-objective evolutionary algorithm based on
  dimension exploration and discrepancy evolution for {UAV} path planning
  problem}.
\newblock \bibinfo{journal}{Information Sciences} \bibinfo{volume}{657},
  \bibinfo{pages}{119977}.
%Type = Inproceedings
\bibitem[{Xue(2017)}]{xue2017uav}
\bibinfo{author}{Xue, M.}, \bibinfo{year}{2017}.
\newblock \bibinfo{title}{{UAV} trajectory modeling using neural networks}, in:
  \bibinfo{booktitle}{17th AIAA Aviation Technology, Integration, and
  Operations Conference}, \bibinfo{organization}{ARC}. p.
  \bibinfo{pages}{3072}.

\end{thebibliography}

%% else use the following coding to input the bibitems directly in the
%% TeX file.

%%%\begin{thebibliography}{00}

%% \bibitem[Author(year)]{label}
%% Text of bibliographic item

%%%\bibitem[()]{}

%%%\end{thebibliography}
\end{document}